\documentclass[pdflatex,sn-mathphys-num]{sn-jnl}

\usepackage{graphicx}%
\usepackage{multirow}%
\usepackage{amsmath,amssymb,amsfonts}%
\usepackage{amsthm}%
\usepackage[title]{appendix}%
\usepackage{xcolor}%
\usepackage{textcomp}%
\usepackage{manyfoot}%
\usepackage{booktabs}%
\usepackage{algorithm}%
\usepackage{algorithmicx}%
\usepackage{algpseudocode}%
\usepackage{listings}%
\usepackage{url}%
\usepackage{bm}%
\usepackage{nicefrac}%
\usepackage{microtype}%
\usepackage{soul}%

\hypersetup{bookmarksdepth=3}

\begin{document}

\title[Detecting Speculative Language in Biomedical Texts]{Detecting Speculative Language in Biomedical Texts using Recurrent Neural Tensor Networks}

\author*[1]{\fnm{Dhruv} \sur{Dixit}}\email{ddixit1@stevens.edu}

\affil*[1]{\orgdiv{Department of Electrical and Computer Engineering}, \orgname{Stevens Institute of Technology}, \orgaddress{\city{Hoboken}, \state{New Jersey}, \country{United States}}}

\abstract{In this investigation, we delve into the automated detection of speculative language within biomedical articles by utilizing distributed sentence representations and advanced deep learning techniques. The implications of such identification extend to information retrieval, multi-document summarization, and the exploration of new knowledge. Our exploration encompasses two distinct approaches for acquiring distributed sentence representations: the Paragraph Vector model and the Recursive Neural Tensor Network. These methodologies are then rigorously compared against three foundational baseline algorithms: Support Vector Machines, Naive Bayes, and pattern matching. Our findings reveal that the Recursive Neural Tensor Network (RNTN) demonstrates a slight performance edge (F1 = 0.885) over the top-performing baseline, the linear bigram SVM (F1 = 0.881). Meanwhile, the Paragraph Vector model proves less effective (F1 = 0.368), even after extensive training using an expansive, unlabeled dataset. We engage in a comprehensive discourse on the factors influencing these performance disparities and provide insightful recommendations for future research directions.}

\keywords{speculative language detection, biomedical text mining, recursive neural tensor network, paragraph vector, support vector machine}

\maketitle

\section{Introduction}
Published articles in the biomedical field include both established solid facts and new emerging less solid knowledge that represents new thoughts or findings \cite{Medlock2008}. Consider the following example sentences:
{\it The study proves that the promoter is effective.}
{\it The study suggests that the promoter may be effective.}
The second sentence includes cue words and phrases such as {\it suggests that} and {\it may be} which render the scope, {\it the promoter is effective} to be speculative. Automatic recognition of such speculative statements would ensure that an Information rretrieval system does not misrepresent less solid information as being facts. such work previously was done using pattern matching \cite{Malhotra2013}, probabilistic weakly supervised learning models and Support Vector Machines \cite{Medlock2007}.  Such methods use the bag of words approach, representing the text, every element indicating or representing word ffrequency  (for unigram BoW) or a pair of words (for bigram BoW). This approach approach has 2 setbacks: it loses word order and also ignores words semantics \cite{Le2014}. The ordering and semantics of the words are important for the speculative language recognition task as the speculativeness of a sentence depends not just on the presence of cue words but also on the context in which they appear.

It has been shown that a distributed representation of words can capture semantic information \cite{mikolov2013efficient}. This is a representation with continuous rather than Boolean or integer features as is the case for BoW. Recent advances in machine learning have made learning the distributed representations of sentences and text of arbitrary length possible as well. this study demonstrates the possibility to using distributeed representation using two approaches: Paragraph Vectors \cite{Le2014} and Recursive Neural Tensor Network \cite{Socher2013}. We hypothesize that applying distributed representation of sentences to the task of speculative language recognition in biomedical articles improves  performance , in comparision to baseline algorithms.  performance from these two methods and that of the baselines is evaluated using their F1 scores, ROC curves and Precision-Recall curves.

Hereon, section two exhibits related work. Section three describes in detail Paragraph Vector representation and Recursive Neural Tensor Network methods. The evaluation of these methods against the baselines considered is presented in Section 4. Section 5 consists of discussions, findings, limitations of the approaches used . We conclude in Section six.

Our code is available online.\footnote{
\url{https://github.com/boonjiashen/speculative-language-recognizer}}

\section{The Related work}
Speculative language recognition, especially in the biomedical domain, has been researched upon for many years. An SVM-based text classification approach and a substring matching approach are discussed in one of the earliest papers addressing this task \cite{Light2004}. In \cite{Medlock2007}, a weakly supervised probabilistic model for training data acquisition and classification is proposed. Their model  uses Naive Bayes with smoothing to avoid unreliable frequency estimates with limited data to train. We explore a similar approach as one of our baselines.We introduce a Bayesian logistic regression classifier that employs a Laplace prior in \cite{vlachos2010detecting}.They tackle the task of detecting speculation by framing it as a token classification exercise, wherein each individual token is designated as a cue.  They further determine the scope of each cue word. A more recent work described in \cite{Malhotra2013} is a pattern-matchingg model using a dictionary of patterns.

\section{Methodology}

\subsection{Datasets}
Our labeled dataset comes from the BioScope corpus \cite{bioscope}; our unlabeled dataset comes from BioMed Central \cite{biomed}. The BioScope corpus is a dataset of biomedical articles described in XML. Each sentence is parsed as follows. For each speculative sentence, the cue word/phrase that causes the speculation is tagged, as well as the scope of speculation. In the following illustrative example,

\begin{verbatim}
<sentence> 
    The novel enhancer element identified in this study is 
    <xcope> 
        <cue type="speculation">probably</cue> 
        a target site for both positive and negative factors.
    </xcope>
</sentence>
\end{verbatim}

 scope of the entire sentence is demarcated,  speculative cue word {\it probably} has been tagged, and the scope of speculation has been tagged as well.

The BioMed corpus also comprises biomedical articles in XML. It differs from the BioScope corpus in two key aspects. It is much larger (over 200000 articles as compared to the BioScope's 1273 paragraph-abstracts and nine studies ), and sentences  not labeled speculative/non-speculative. We utilized BioMed corpus' subset for upscaling Paragraph Vector approach training. All other approaches are trained and tested on the BioScope corpus only.

All sentences in the dataset are converted to lower-case.

\subsection{Paragraph vector representation}

The paragraph vector (PV) approach follows that described in \cite{Le2014}. It comprises two models. At test time, the paragraph vector model outputs a fixed length vector for the test sentence. Then a logistic classifier outputs a binary label given the paragraph vector.\footnote{To remain consistent with the terms used in \cite{Le2014}, we use the term `paragraph vector' to refer to the fixed length feature vector of a sentence. In this study, test and training instances are sentences, but the paragraph vector algorithm is able to learn feature vectors for text of arbitrary length (e.g., phrases, sentences and paragraphs).}

\paragraph{Dataset} The dataset is partitioned as follows. We assign 70\% of the sentences of the BioScope corpus to traininng seet and 30\% to test dataset.. To exploit a large unlabeled dataset, we append sentences from a subset of the BioMed corpus to the training set, if applicable. As such, the PV model is trained on a mixture of labeled and unlabeled sentences.

\begin{figure}[htbp]
\centering
    \includegraphics[width=0.65\textwidth]{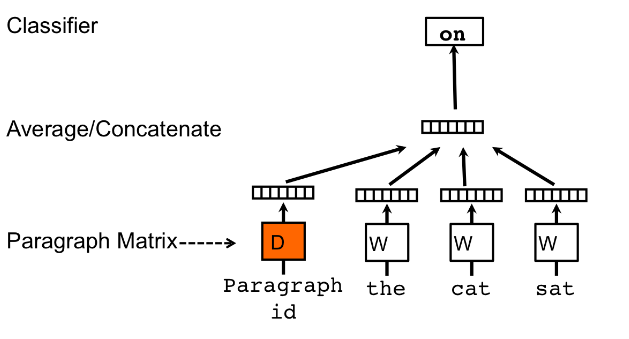}
\caption{Network architecture, Paragraph vector model \cite{Le2014}}
\label{fig:PV}
\end{figure}
\paragraph{Training the PV model}  It is known as the Distributed Memoory approach to Para. Vectors. The model maintains a word vector matrix $W \in \mathbb{R}^{N \times p}$ and a paragraph vector $P \in \mathbb{R}^{M \times p}$, where $N$ refers to amounting training sentences, $M$ stands for unique words of training set, $p$ refers to one feature vector. The input layer is a concatenation of the paragraph vector of the training sentence and the word vectors of the words in a sliding window over the training sentence. output's latyer  is softmax. training objective to predict next word after the sliding window. This objective function is minimized via stochastic gradient descent where the gradient is computed via backpropagation. The error gradients are backpropagated up to and including the input layer.

At training time, we learn $W$, $P$ and the softmax weights. We use the F1 score of a logistic classifier to evaluate the quality of the PV model and to decide when to stop training.\footnote{We do this instead of monitoring the training error due to a limitation in the \texttt{gensim} library. The API does not return the training error after each epoch.} After every training epoch, we learn a logistic classifier with the learned paragraph vectors that have labels. These are the sentences from the BioScope corpus. We stop training the PV model when the F1 score of this logistic classifier starts to drop, i.e., just after the F1 score is maximized. See the discussion section for other evaluation methods that have been considered. The learning rate is constant at 0.001 while the feature length $p$ is 52. No cross-validation is performed.

\paragraph{Training the logistic classifier} After the PV model has been trained, we train the second model of the PV approach, which is a logistic classifier. The training examples are the learned paragraph vectors from the BioScope corpus and their corresponding labels. In practice, we just take the logistic classifier last trained during the training process of the PV model (see above).

\paragraph{Testing a new sentence} At test time, we freeze the softmax weights and $W$, and run a sliding window over the test sentence to learn a paragraph vector by stochastic gradient descent. This paragraph vector is then labeled by the logistic classifier. The number of epochs at test time is held constant at 50 and the learning rate is 0.001.

\paragraph{Implementation details} The PV model approach was implemented in Python. We use the \texttt{gensim} library for the PV model and the \texttt{scikit-learn} library for the logistic classifier. To split a textblock from the BioMed corpus to a list of sentences, we use the NLTK English Punkt Sentence Tokenizer. To split a sentence into a list of words/punctuation, we use NLTK's \texttt{nltk.word\_tokenize} function.

\begin{figure}[htbp]
\centering
\includegraphics[width=6.5cm,height=8.2cm,keepaspectratio]{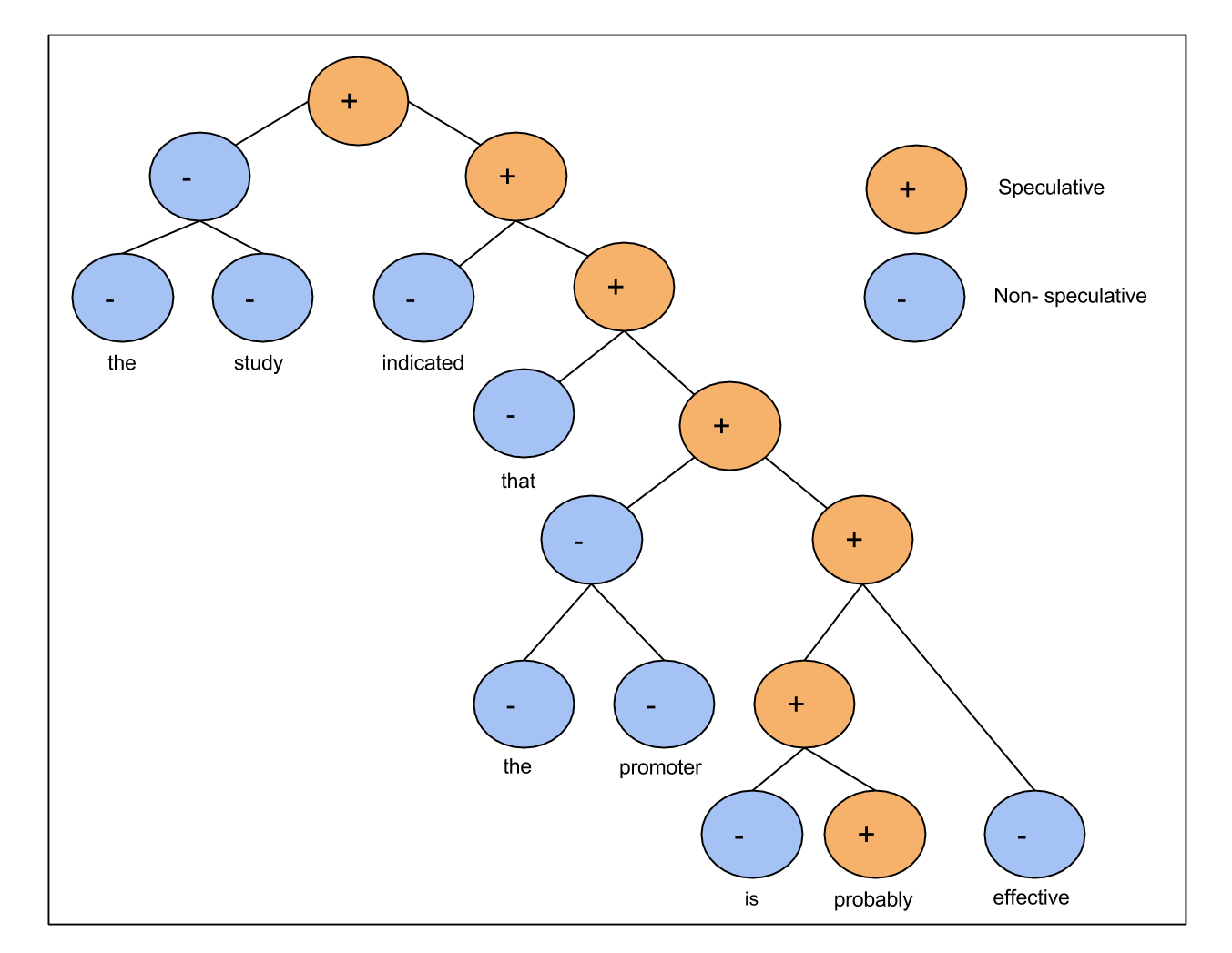}
\caption{Speculation labeled parse tree for the example sentence {\it the study indicated that the promoter is probably effective}}
\label{fig:parsetree}
\end{figure}

\subsection{Recursive Neural Tensor Networks}

The limitations of a semantic vector space confined to individual words become evident when attempting to adequately capture the intricate meanings embedded in longer phrases. To counteract this shortfall, \cite{Socher2013} put forth an innovative strategy that involves harnessing fully annotated parse trees to dissect the nuanced composition of sentiment embedded within language. The Stanford Sentiment Treebank, which serves as the cornerstone of this research, encompasses binary parse trees generated using the Stanford parser and meticulously annotated by human evaluators. Seeking to enhance the precision of compositional analysis, \cite{Socher2013} introduced the Recursive Neural Tensor Network (RNTN). This groundbreaking neural architecture accommodates phrases of varying lengths as input data. It adeptly represents phrases through the amalgamation of word vectors and parse trees, subsequently deriving vectors for higher-level nodes within the parse tree by applying a consistent tensor-based composition function. We explore possibility tp apply the same idea to the task of speculative language recognition.  Since our data set - the BioScope corpus \cite{Vincze2008} has the speculative cue words tagged, we use it for annotating our parse trees instead of human judges.

We now describe the generation and labeling of parse trees and training the RNTN with an example sentence: {\it``the study indicated that the promoter is probably effective''}.

\begin{figure}[htbp]
\centering
\begin{minipage}{.45\textwidth}
\centering
\includegraphics[width=6.5cm,height=8.2cm,keepaspectratio]{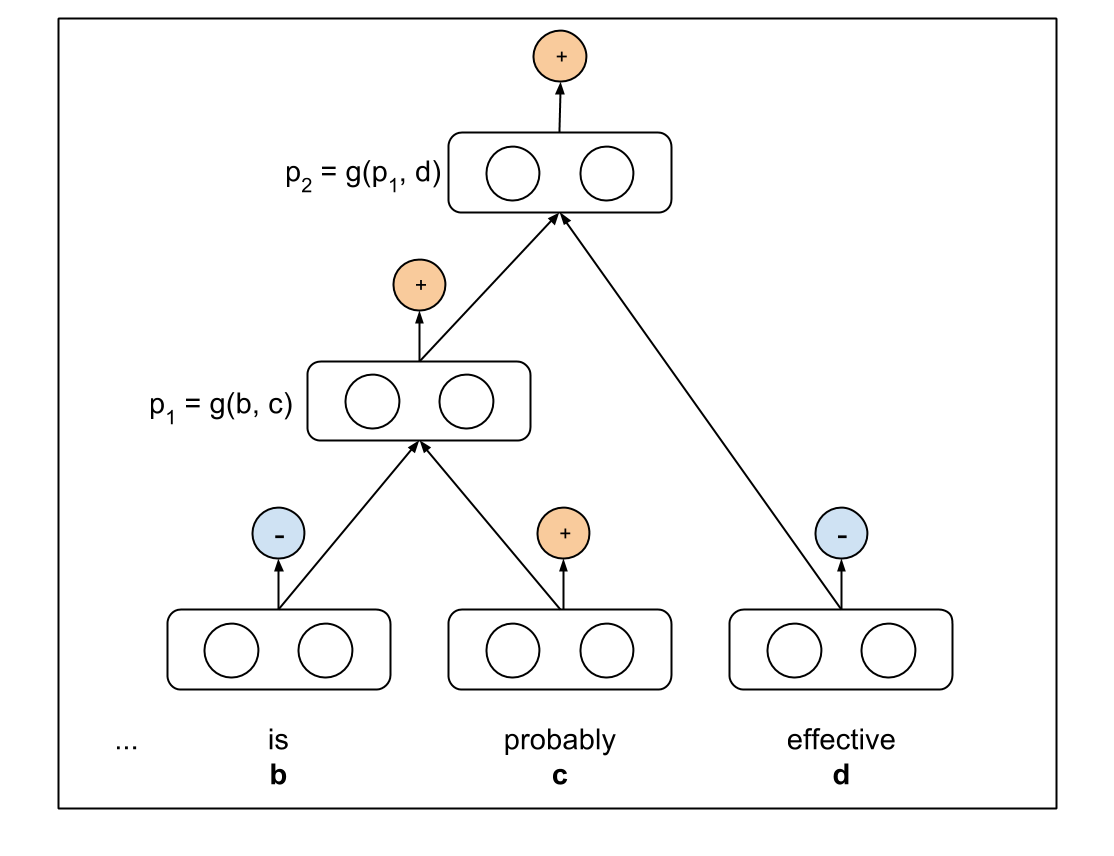}
\caption{Generate higher-level vectors through an ascending procedure utilizing a compositionality function labeled as g, and employ node vectors as distinct characteristics for a classifier positioned at the corresponding node\cite{Socher2013}}
\label{fig:RNTNexpl}
\end{minipage}
\begin{minipage}{.1\textwidth}
\end{minipage}
\begin{minipage}{.45\textwidth}
\centering
\includegraphics[width=6.6cm,height=7.6cm,keepaspectratio]{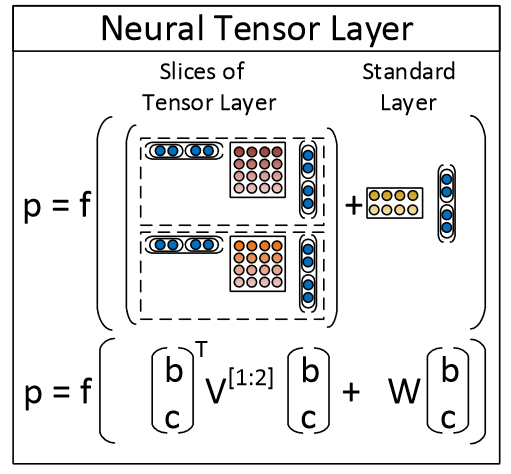}
\caption{Solitary tier of the Recursive Neural Tensor Network, with every dashed compartment depicting one of the {\it d} sections, adept at capturing a particular form of effect that an offspring can impart upon its progenitor.\cite{Socher2013}}
\label{fig:RNTN}
\end{minipage}
\end{figure}

\paragraph{Parse tree generation and labeling} The sentence is first fed into the Stanford parser which tags the Parts-of-Speech (PoS) of the sentence. The nodes of this parse tree are then labeled automatically using the following heuristic: {\it Any node that contains the entire cue phrase is labeled as speculative and all other nodes are labeled as non-speculative}. The labeled parse tree for the example sentence is shown in Figure~\ref{fig:parsetree}. A `$+$' label indicates a speculative node and a `$-$' label indicates a non-speculative node. Such labeled parse trees are obtained for the 14539 sentences taken from the BioScope corpus, which are then split into training/cross-validation/test sets in the proportion 60\%/10\%/30\% respectively.

Tokenizinh of test as well as training sentences is done using the Stanford parsser. This is an intelligent word tokenizer which treats brackets, commas and periods as words. By observation, it seems to parse similarly to the NTLK \texttt{ntlk.word\_tokenize} function used in the PV approach.

\paragraph{Training the RNTN} The labeled parse trees are then trained using a RNTN as described in \cite{Socher2013}. Every individual term is denoted by a matrix with dimensions $d \times d$. The elements of this matrix are adjusted to mitigate the categorization disparity at every vertex. A tensor based composition function is used for all the nodes. In figure four, b' and c' denote the word vectors of the subordinate nodes (refer to Figure~\ref{fig:RNTNexpl}), while $V \in \mathbb{R}^{2d \times 2d \times d}$ represents the tensor defining numerous bilinear structures. The merged vector of b' and c' is subject to multiplication by each segment of $V \in \mathbb{R}^{d \times d}$, and $W \in \mathbb{R}^{d \times 2d}$ embodies the acquired weight matrix.

In conjunction with the word matrices, every node is equipped with a softmax classifier that undergoes training using its vector representation, aiming to forecast a target vector marked as $t$. The target distribution vector for each node is constructed through a 0-1 encoding approach. Since there are two classes for our task, namely {\it speculative} and {\it non-speculative}, $t$ is of length 2 and has a form $[1,0]$ and $[0,1]$ for speculative and non-speculative classes respectively. Utilizing the KL-divergence between the envisioned and the precise outcomes, the objective is to enhance the likelihood of accurate predictions. The propagation of inaccuracies in the RNTN framework is achieved by navigating its intrinsic configuration. For a profound understanding of this procedure, individuals are guided to the exhaustive elaboration in \cite{Socher2013}. 

\paragraph{Testing new sentences}A binary parse treee is developed for every sentecne using the Stanford parser. The trained RNTN model is then used to predict the speculativeness at parse tree's every node. sentence is classified as speculative or not based on the classification of the root node.

\section{Evaluation}

\subsection{Baseline algorithms}

We apply three baseline algorithms to the task of recognizing speculative language in sentences - linear Support Vector Machines (SVM), Naive Bayes (NB) and pattern matching.

\paragraph{Linear SVM, NB} We pre-process sentences the same way with linear SVM and NB. Each sentence is transformed into a Boolean Bag-of-Words (BoW) representation.  Each element in the BOW represents the presence or absence\footnote{Note that we only consider whether a unigram/bigram is present or absent, rather than the number of times the unigram/bigram appears in a sentence. We did try the latter approach on linear SVMs and NB, but it consistently performed worse than the Boolean approach. Hence we leave it out of our report.} of a word, in the case of unigrams, or a pair of words, in the case of bigrams. We train the SVM and NB classifiers with 70\% of the dataset and test them on 30\% of the dataset. No cross-validation is done\footnote{We initially tried to cross-validate the penalty term for the linear SVM but found no significant difference in performance between a wide range of values. This strongly suggests that the bigram BoW representation of the training data is linearly separable.}. We use an SVM penalty term $C$ of 1 and a NB Laplacian smoothing term of 1.

\paragraph{Pattern matching} We use a simple, nonparametric method for pattern matching, in which we identify patterns that are considered speculative from the training data and classify a test sentence as speculative if it contains any of these patterns. The patterns identified as speculative are the phrases annotated as speculative cues in the Bioscope corpus. We use a train-test split of 70\%-30\% as before, with no cross-validation.

\section{How to evaluate the PV model}
One drawback of an unsupervised learning method such as the Paragraph Vector model is the difficulty in evaluating the learned model. In this study, we considered two evaluation methods that evaluated only the learned word vectors, based on Google's word relationship set\footnote{\url{https://code.google.com/p/word2vec/}} \cite{mikolov2013efficient}. Each relationship is a 4-tuple that can be categorized as a semantic relationship (e.g., Italy is to Rome as France is to Paris) or a syntactic relationship (e.g. run is to running as grow is to growing). With this 4-tuple and word-to-vector mapping, we can try to predict one of the words in the 4-tuple given the other three words, e.g., try to answer `Paris' given the question `{\it Italy} is to {\it Rome} as {\it France} is to {\it ?}'.

One possible evaluation metric would be the accuracy of a prediction task, in which a model is considered correct if the true label (e.g., Paris) is in one of the model's top 5 predictions.  Another possible metric is the average distance between the two deltas of the 4-tuple, e.g. we measure the distance between $vec(Italy) - vec(Rome)$ and $vec(France) - vec(Paris)$, where $vec$ is a function that maps a word to its vector.  A good model should give a small distance between these two terms. The metric is the average distance of the set of 4-tuples.

To this end, a better word relationship evaluation set should be developed specific to biomedical language. In Google's relationship set, much of the vocabulary in the semantic relationships is absent in biomedical language. For example, you will be hard-pressed to find the names of countries and capitals in biomedical articles. Hence, a large portion of the relationship set has to be excluded if used on a biomedical dataset. We suggest creating a set of semantic relationships for biomedical language (e.g., headache is to aspirin as cough is to dextromethorphan). While these may not be as unequivocal as country-capital relationships, they serve as a quick way to evaluate word vectors.

\subsection{Empirical Evaluation}
\begin{table}[htbp]
\caption{Performance metrics ranked by F1 score}
\centering
\begin{tabular}{|l c c c c|}
\hline
&&&&\\
\textbf{Method} & \textbf{Accuracy} & \textbf{Precision} & \textbf{Recall} & \textbf{F1 score} \\
\hline
&&&& \\
PV (Bioscope + 1000 BioMed articles) & 0.761 & 0.341 & 0.368 & 0.354 \\
PV (BioScope only) & 0.704 & 0.291 & 0.463 & 0.357 \\
PV (Bioscope + 3000 BioMed articles) & 0.735 & 0.320 & 0.433 & 0.368 \\
Bigram NB & 0.900 & 0.909 & 0.479 & 0.627 \\
Pattern Matching & 0.805 & 0.468 & 0.985 & 0.635 \\
Unigram NB & 0.903 & 0.738 & 0.694 & 0.715 \\
Unigram SVM & 0.957 & 0.911 & 0.835 & 0.871 \\
Bigram SVM & 0.962 & 0.964 & 0.811 & 0.881 \\
RNTN & 0.958 & 0.904 & 0.866 & 0.885 \\
\hline
\end{tabular}
\label{table:F1score}
\end{table}

\begin{figure}[htbp]
\centering
\includegraphics[width=10cm,height=9cm,keepaspectratio]{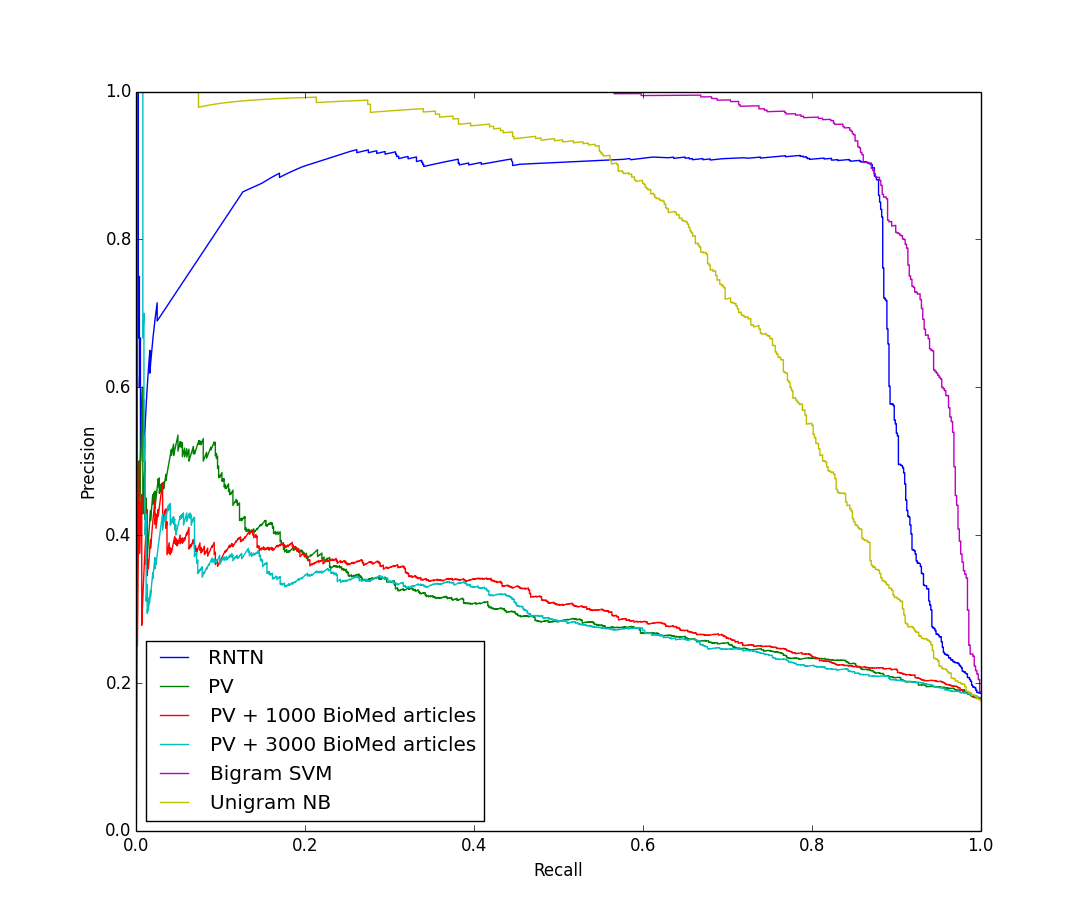}
\caption{Precision-recall curves for PV approach, RNTN and baselines.}
\label{fig:PR}
\end{figure}

\begin{figure}
\centering   \includegraphics[width=10cm,height=9cm,keepaspectratio]{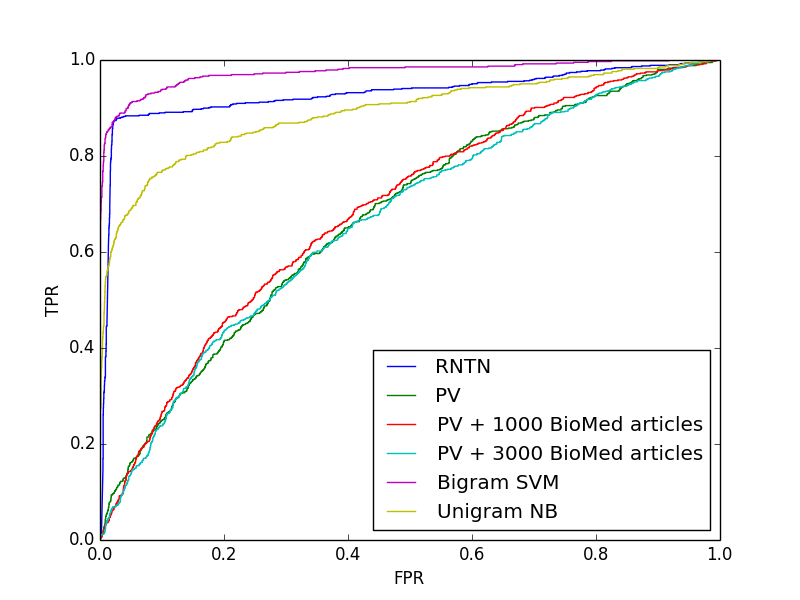}
\caption{ROC curves for PV approach, RNTN and baselines}
\label{fig:ROC}
\end{figure}

A summary of the results are given in Table~\ref{table:F1score}. Precision-recall curves and ROC curves are shown in Figures~\ref{fig:PR} and \ref{fig:ROC} respectively. Bigram NB and unigram SVM do not perform as well as unigram NB and bigram SVM respectively and are not shown in the figures.

The RNTN has the best F1 score of 0.885, performing marginally better than bigram SVMs (F1 = 0.881). The PV model with no unlabeled data performs the worst (F1 = 0.357), and performance does not improve even after training with 1000 BioMed articles (approx 370K sentences) and 3000 BioMed articles (approx. 1.1 million sentences).


\section{Discussion}

\paragraph{Baselines} The linear bigram SVM model performs significantly better than both unigram and bigram NB, which suggests that the feature independence assumption of Naive Bayes does not hold true in our data. Due to the nature of pattern matching, the algorithm has very high recall and low precision. High recall occurs because most of the speculative cues that exist in the test set appear in the training set. Low precision occurs because speculative cues may not be speculative depending on their context, e.g., the cue {\it or}. Therefore a speculative cue that appears in a training sentence and appears again in a test sentence may not be speculative in the latter case, resulting in a false postive.

\paragraph{PV approach} There may be several explanations to why the PV approach does poorly. One is that the model failed to converge to a local minimum as the F1 score of the logistic classifier is a poor surrogate for the training accuracy of the PV model. Another is that the training data cannot express the semantic information of its constituent sentences, and therefore a learner cannot learn this information given the data. Intuitively, it would be difficult even for a native English speaker to learn medical biology just by reading biomedical articles, since these articles assume an expert level of knowledge in the field. A similar scenario may be occurring in this unsupervised training algorithm. For this reason, it may prove useful to train the model initially with easier text such as college-level textbooks before fine-tuning the model with academic articles. Yet another possible explanation is that the model has overfit the training data, and should be trained with more data. Note that we only used about 1\% of the articles in the BioMed corpus in this study.

\paragraph{RNTN} The RNTN approach produces much better results than the Paragraph Vector Representation, suggesting that speculative language recognition is amenable to formulation as a sentiment analysis task using RNTN.

The ROC and precision-recall curves for RNTN have a more angular structure than the other methods. This is because the confidence value for a sentence identified as speculative is very close to 1, causing sharp angles in the beginning and the ends of the ROC and precision-recall curves respectively. In other words, when the classifier identifies a sentence as speculative, it does so with high confidence,  regardless of correctness.

While RNTN produces the best F1 score in our experiments, its performance is just marginally better than our best performing baseline, linear bigram SVMs. The comparable success of linear SVMs indicates that  speculative language recognition may be reasonably approximated as a classification task for linearly separable data, and may not benefit significantly from more sophisticated approaches using distributed word representations. Additionally, one disadvantage of RNTN is that its training is significantly slower than the other methods.

\subsection{Limitations of our approach} 
The features used by our methods do not include meta-information such as author and affiliations, which may have a prior on propensity for speculative language. This information if available, can be used for validating the reliability of the information in a biomedical article. Also, each sentence is considered as an independent sentence and does not consider neighboring sentences or paragraphs. Furthermore, in our implementations, we have considered the absence of a speculative cue word/phrase as indicative of a non-speculative sentence. But there are cases when a sentence without a speculative cue word has some speculation.As an example, contemplate a declaration concerning an outcome that may not inherently result from the work disclosed, yet has the potential for extrapolation \cite{Light2004}. Even though this assertion might not include speculative cue words, it should not be regarded as an established fact.

A technical limitation of the PV approach is that it stores the paragraph vectors of all training sentences. Therefore its memory complexity is linear in the number of training sentences, compared to RNTN which is linear in the size of the vocabulary but approximately constant in the number of training sentences. Of the two, RNTN scales better with larger amounts of training data. However, RNTN is trained only on labeled data, so it cannot exploit the large set of unlabeled biomedical articles.

\section{Conclusion}

We investigate the application of the distributed representation of sentences to the task of speculative language recognition in biomedical text. We try two methods that transform a test sentence into a distributed representation, RNTN and the PV approach, and compare these two methods against several simple baseline algorithms. We find some evidence to support our hypothesis that this representation improves performance in the recognition task. The RNTN performs slightly better than linear bigram SVM, our best-performing baseline algorithm, but we also note that the marginal improvement of 0.4\% in F1 score is at the cost of high computational time. The PV approach performs significantly worse than all other algorithms in this study, even when the PV learner is trained with a large unlabeled dataset. 

The PV approach has the potential to learn high-quality distributed representations of sentences by exploiting a large unlabeled dataset. It is the only algorithm in this study that can be unsupervised. However, how to tune the parameters used at test time is an open problem. Future research may include investigating this problem, or pretraining the model with general text such as Wikipedia articles. In addition to classifying sentences, we may also investigate determining the scope of the speculative cue words in speculative sentences.

\backmatter

\bmhead{Code availability}
The code associated with this study is available at \url{https://github.com/boonjiashen/speculative-language-recognizer}.

\bibliography{bibliography}

\end{document}